\documentclass{bmvc2k}

\title{Pose-Transformation and Radial Distance Clustering for Unsupervised Person Re-identification}

\addauthor{Siddharth Seth}{ssiddharth@iisc.ac.in}{1}
\addauthor{Akash Sonth}{akashsonth@vt.edu}{1}
\addauthor{Anirban Chakraborty}{anirban@iisc.ac.in}{1}

\addinstitution{
 Visual Computing Lab,\\
 Department of Computational and Data Sciences,\\
 Indian Institute of Science,\\
 Bangalore, India
}

\runninghead{Seth, Sonth, Chakraborty}{Unsupervised Person Re-id}

\begin{document}

\maketitle

\begin{abstract}
Person re-identification (re-ID) aims to tackle the problem of matching identities across non-overlapping cameras. Supervised approaches require identity information that may be difficult to obtain and are inherently biased towards the dataset they are trained on, making them unscalable across domains. 
To overcome these challenges, we propose an unsupervised approach to the person re-ID setup. Having zero knowledge of true labels, our proposed method enhances the discriminating ability of the learned features via a novel two-stage training strategy. The first stage involves training a deep network on an expertly designed pose-transformed dataset obtained by generating multiple perturbations for each original image in the pose space.
Next, the network learns to map similar features closer in the feature space using the proposed discriminative clustering algorithm. We introduce a novel radial distance loss, that attends to the fundamental aspects of feature learning - compact clusters with low intra-cluster and high inter-cluster variation. Extensive experiments on several large-scale re-ID datasets demonstrate the superiority of our method compared to state-of-the-art approaches.
\end{abstract}


\section{Introduction}
\label{sec:intro}

Person re-identification deals with matching the identity of a query image from a dataset consisting of person images taken across a disjoint set of cameras. There has been considerable research on this problem using supervised \cite{zhang2020relation, zhuang_eccv20, tay_cvpr19, chen_iccv19, zheng2019joint, ge2018fd}, one-shot \cite{wu_cvpr18, ye_eccv18}, transfer learning \cite{deng_cvpr18, hehe_18, peng_cvpr16}, unsupervised domain adaptation \cite{Jin_2020_CVPR, ge2020mutual, zhai2020multiple, zhong2019invariance, zhai_cvpr20, Li_2019_ICCV, huang_iccv19}, and unsupervised learning \cite{liao_cvpr15, lin2019bottom, ding_bmvc19, wu_aaai20, lin_cvpr20, wang_cvpr20} frameworks. Supervised approaches require a labeled dataset having both identity and camera information, making them inherently biased towards the dataset they are trained on and are therefore not generalizable. The unsupervised counterpart does not assume the availability of any such labeled information making it more deployment friendly in real-world scenarios.

\textit{Unsupervised domain adaptation} (UDA) methods \cite{Jin_2020_CVPR, ge2020mutual, zhai2020multiple, zhong2019invariance, zhai_cvpr20, Li_2019_ICCV, huang_iccv19} tackle the problem of domain generalizability by training on some available labeled source dataset and then adapting to the target unlabeled dataset. Though such methods have shown to yield state-of-the-art results, they raise two important concerns: 1) labeled dataset is expensive to obtain, restricting the scalability to real-world environments, and 2) there is no way to ascertain the domain gap between the source and target datasets. Thus, one cannot ascertain the ideal dataset that will yield the most superior domain transfer on target. \textit{Purely unsupervised} methods \cite{liao_cvpr15, lin2019bottom, ding_bmvc19, wu_aaai20, lin_cvpr20, wang_cvpr20} differ from UDA techniques as they delve one step further into the complexity of the problem by being completely unaware of any identity information, either in source or target dataset. In this work, we propose a novel discriminative clustering approach for the purely unsupervised re-ID setup.

Learning without labels is challenging for any problem in general. For the re-ID task, this involves learning discriminative features from the unlabeled dataset to identify query images in the gallery set. Initially, we create pseudo labels by assigning each image its own unique label. However, a simple re-ID training using pseudo labels as the supervisory signal fails to learn any useful features. We thus seek to enhance the network’s discriminative ability using a two-stage training procedure. We first take each image and generate similar looking, pose-transformed images in the identity space. These are then assigned labels corresponding to their original images. This helps project the image identity into different poses, thereby yielding a set of more realistic augmentations, thus giving the model an insight into different camera angles inherently present in the dataset. The base CNN 
is then trained on these samples using the aforementioned pseudo labels as supervision. This is done in conjunction with a metric learning loss to learn a latent space where data points belonging to different identities are farther apart than those belonging to the same identity.

The second stage builds on the first stage by using the proposed discriminative clustering algorithm.
Plugging out the pose-transformed samples from the dataset, this step only lets the network access the original unlabeled dataset. This not only eases the burden of training with a large dataset but also helps in learning true distribution features. Our novel loss function aims to explicitly move points in different clusters farther apart and the points inside a cluster within a certain distance, thereby increasing the inter-cluster variation and decreasing the intra-cluster variation. This helps the network achieve powerful discriminability.

Our contributions are summarized as follows:
\begin{itemize}

\item We design a pose-transformed dataset that does not require any identity information, is non-parametric, and can be adopted to any person re-identification method.
%

\item We introduce a novel \textit{radial distance} loss that explicitly attends to inter and intra cluster variations by maintaining a minimum \textit{radial distance} between clusters while pushing dissimilar samples away. We also show that it helps discerning similar looking identities.

\item We achieve state-of-the-art results for the unsupervised person re-identification task on the  Market-1501 \cite{zheng2015scalable}, DukeMTMC-reID \cite{zheng_iccv17}, and MSMT17 \cite{wei2018person} datasets.
\end{itemize}

%


\section{Related Works}
\label{sec:related}
Traditional approaches using hand-crafted features \cite{lisanti_tpami15, liao_cvpr15, Farenzena_cvpr10, bingpeng_bmvc12, bingpeng_eccvw12}, dictionary learning \cite{kodirov_bmvc15}, and saliency analysis \cite{zhao_tpami17, wang_bmvc14} 
did not implicitly extract relevant features invariant to multiple factors of variations (viewpoint, illumination, background clutter).

\textbf{Unsupervised Person re-ID.} 
Unlike traditional approaches, deep methods \cite{wang_cvpr20, lin2019bottom, ding_bmvc19, lin_cvpr20} instead give the network the freedom to learn discriminative features. \cite{zhuang_eccv20} use camera-based batch normalization that disassembles re-ID datasets and aligns each camera independently.
\cite{lin_cvpr20} propose a cross-camera encouragement term that relieves the issue of camera variance. However, intra-camera identity or camera IDs knowledge is essential for such approaches, which may not always be feasible.
\cite{wang_cvpr20} use multi-label classification for label prediction and a memory bank to store the updated image features. %
\cite{Wang2021camawareproxies} assign camera-aware proxies for dealing with intra-ID variance, thus assuming camera ID information.
\cite{wu_aaai20} use tracklet data
thereby assuming the availability of temporal information. In light of the above observation, we formulate the solution to the unsupervised person re-ID problem in the simplest form with no assumption of any form of identity related supervision.

\textbf{Generative approaches.}
With the emerging popularity of generative adversarial networks, several recent works \cite{ge2018fd, huang_tip18, liu_cvpr18, xuelin_eccv18, zheng_iccv17, zheng2019joint, NIPS2019_8771, yang_eccv20} adopt a procedure to augment the dataset by generating realistic looking training samples.
The generative module in \cite{zheng2019joint} generates high quality cross-id images that are used to augment the dataset online. \cite{Li_2019_ICCV} disentangle pose and appearance information using a labeled source dataset and adapt to target dataset. However, all the above methods are either supervised in the target domain or in the source domain for UDA approaches and are therefore unsuitable in an unsupervised person re-ID scenario. On the contrary, we present a simple method for generating image-level samples by spatially transforming the images without leveraging any image identity information, making the approach completely unsupervised.

\textbf{Clustering-based methods.}
Clustering algorithms have been used to identify and group similar features together for better discriminability. \cite{wang_cvpr20, lin2019bottom, ding_bmvc19} initially assign each image to belong to its own cluster and then progressively merge clusters. \cite{lin2019bottom} use bottom up clustering to automatically merge similar images into one cluster. However, they make a strong assumption that images are evenly distributed between the identities
which will almost never be true for any dataset, and will fail to work when deployed to a real world scenario. \cite{ding_bmvc19} explicitly take care of cluster compactness and inter-cluster separability using dispersion based clustering
by measuring the distance between all members between two clusters. \cite{wang_cvpr20} propose to use multilabel classification to identify similarity between images. The method inherently optimizes the inter- and intra- class distances.
However, simply measuring the distance between two points does not properly ensure cluster separability and compactness.

\textbf{Unsupervised learning.} There are several works that focus on learning discriminative features. Center loss \cite{wen_eccv16}, for instance, tries to make every cluster compact by projecting the cluster features at the center. DeepCluster \cite{caron2018deep} and SwAV \cite{caron2020unsupervised} have been shown to be effective for image classification tasks where both shape and appearance of an object play a role, such as in large-scale datasets like ImageNet \cite{deng2009imagenet}. Recently, \cite{Geirhos2019a} highlight that reducing bias of deep networks on texture helps improve ImageNet accuracy. However, learning discriminative features for person re-ID requires the model to discern between different identities of humans that frequently appear same due to similar apparel and/or occlusion. Though such unsupervised techniques still work well for re-ID, they often do not explicitly tackle the problem of inter and intra cluster variation. Those that do \cite{wen_eccv16, ding_bmvc19, wang_cvpr20}, fail to address the challenging problem of distinguishing between similar looking identities.

Different from such works, we use clustering with the proposed \textit{radial distance} loss that addresses the twin goals of 1) maintainging high inter and low intra cluster variation, and 2) discerning similar appearing identities.


\begin{figure*}
\begin{center}
	\includegraphics[width=1.0\linewidth]{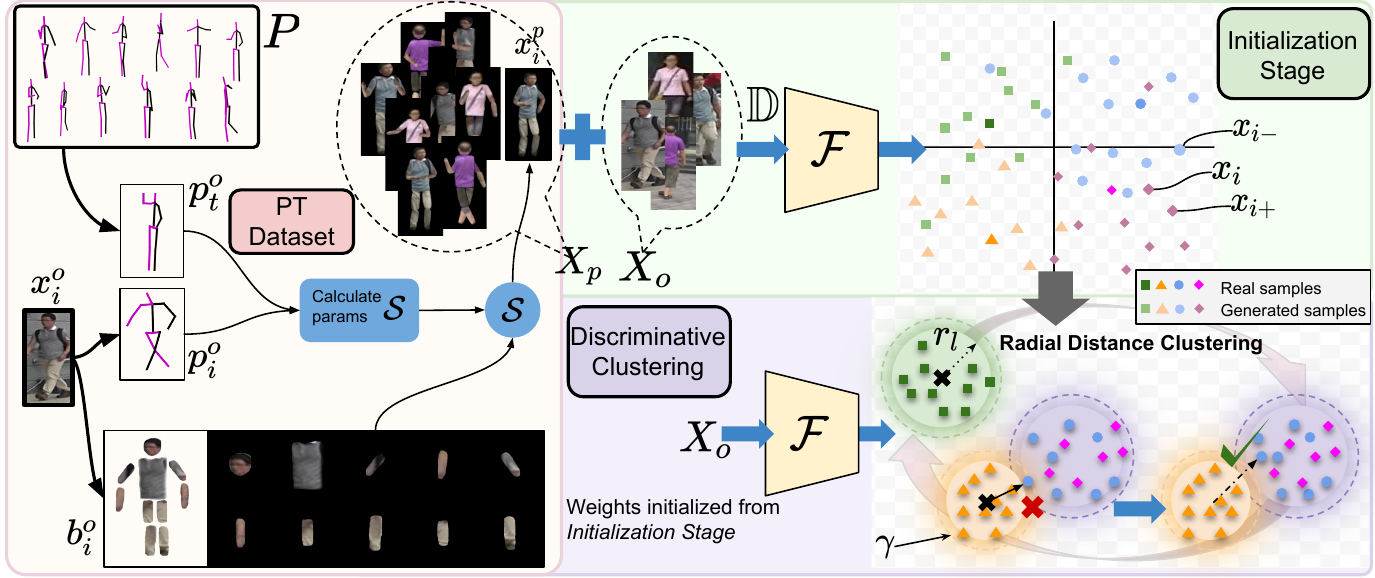}
	\caption{ 
	Pipeline of the proposed person re-id framework. PT Dataset: Poses and parts corresponding to the original dataset are obtained. Parts are transformed spatially as per the sampled poses, $p^o_t$ to generate images, $x_i^p$. PT dataset $=$ original dataset $+$ generated samples is used to train the initialization stage. Initialization Stage: Trained using triplet and classification losses. Discriminative Clustering: Weights for $\mathcal{F}$ are taken from the Initialization Stage. $\mathcal{F}$ is trained only on the original dataset using triplet, classification losses and the proposed radial distance loss. $\gamma$ represents the minimum radial distance. The blue point violates $\gamma$ for Orange cluster, and thus, is pushed away.
	}
 	\label{fig:side2}    
\end{center}
\end{figure*}

\section{Approach}
\label{sec:approach}
In this section, we describe our proposed unsupervised learning framework for person re-ID. Let the training set be denoted as ${X}_{o} = \{x_i^o\}_{i=1}^N$, where $x_i^o \in \mathbb{R}^{H\times W\times3}$ with no associated labels. Our goal is to learn a feature representation using a network $\mathcal{F}$ on this unlabeled dataset. Initially each image is assigned its own unique label. The set of pseudo labels are then defined as $Y = \{y_i^o\}_{i=1}^N$ where $y_i^o \in \{1,2,..,N\}$. We create an auxiliary dataset $X_p = \{x_j^p\}_{j=1}^{K\times N}$, where $K$ is a constant integer scaling factor.
The set $X_p$ is constructed via $K$ different pose transformations applied on each image in $X_o$, where $K$ is a hyperparameter. For each $x_i^o$, the corresponding $K$ pose-transformed images are assigned the label $y_i^o$. The pseudo labeled Pose-Transformed (PT) dataset thus becomes $\mathbb{D} = \{X_o	\cup X_p,\;Z\}$, where $Z = \{y_i\}_{i=1}^{(K+1)N}$ and $y_i \in \{1,2,..,N\}$. For simplicity and future references, we denote the samples in the Pose-Transformed dataset $\mathbb{D}$ as $\{x_i,\: y_i\}$.
The pseudo labels, thus defined, enable us to transform the unsupervised re-ID setting into a supervised one. We initially train a deep network $\mathcal{F}$ on $\mathbb{D}$ in a supervised fashion. However, unlike the trivial augmentation functions, such as flip, tilt, random erase, etc., the pose transformations result in a much larger diversity for the data samples per class, akin to same exhibited by the target images observed across a camera network. As a result, the proposed PT dataset can be utilized to provide a strong supervision to the network and lay the groundwork for a better discriminative training.

Next, we seek to improve the discriminative power of $\mathcal{F}$ by further training it via clustering and the proposed \textit{radial distance} loss. The aim is to pull similar features closer and push dissimilar features further apart, the fundamental goal of any metric learning framework. %
Evaluation is performed by measuring the Euclidean distance between the extracted features of query and gallery images using $\mathcal{F}$.

\subsection{Pose-Transformed dataset}
Pose-Transformed (PT) dataset is modeled by spatially transforming the body parts of a person in a re-ID image to a different random pose. We choose to transform each image into \textit{K} different poses. To achieve this, we extract the poses, $P = \{p_i^o\}_{i=1}^N \in \mathbb{R}^{15\times3}$ corresponding to each person in $X_o$ using the pretrained pose estimation model \cite{kolotouros2019spin}. The aim now is to transform the person in image $x_i^o$ from pose $p_i^o$, to a different randomly sampled pose, $p^o_t$, $t \neq i$ from the set of poses $P$. We utilize the same set of poses extracted from the target images to ensure that PT dataset adheres to a re-ID setup.

\textbf{Body part masks.} We propose to use two techniques for extracting body parts, represented by $b_i^o \in \mathbb{R}^{10\times H\times W\times3}$, 1) using the aforementioned pose estimation method \cite{kolotouros2019spin}, and 2) manually annotating the body parts of a \textit{single} image and utilizing them throughout the dataset. The latter has the advantage of not being restricted to use a method that estimates human poses and body parts simultaneously. Through a quantitative comparison in Sec~\ref{sec:ablation}, we
show that the overall strategy is robust to using any of these techniques.

\textbf{Parts transformation.}
The final step performs the alignment of the detected body parts to the randomly sampled target pose, $p^o_t$. This is done by computing the parameters (rotation, scaling and translation) of the spatial transformation, $\mathcal{S}_t$, required for transforming the detected pose, $p_i^o$ to the target pose, $p^o_t$. We then get the transformed sample as, $x_i^p = \mathcal{S}_t \circ b_i^o$.

Our aim here is to learn representative features from the dataset for each identity. During spatial transformation, minute details from the face and body accessories may get distorted or removed. But most importantly, we retain the body texture from the original image. This is unlike the deep generative models \cite{tang2020xinggan, ma2017pose, ma2018disentangled} that attempt to reconstruct a person in different poses. Even though they successfully reconstruct the person, the texture is comparatively different than the original image. We thus seek to generate ``a set of perturbations for each original target'', that does not focus solely on image quality but on retaining most of the features present in the original image. This is particularly useful for the re-ID task where two different identities may look quite similar. Please refer Supp. for more details.%
\subsection{Initialization stage}

The objective of the initialization stage is to build a groundwork for learning strong discriminative features in the discriminative clustering stage.

\textbf{Re-identification objective.} Using the aforementioned pseudo labeled dataset, $\mathbb{D}$, the network $\mathcal{F}$ optimizes for identity classification using a cross entropy loss,

\begin{equation} \label{eq1}
\mathcal{L}_\textit{cls} = \frac{1}{(K+1)N} \sum_{i=1}^{(K+1)N} \log p(y_i| x_i) 
\end{equation}

\textbf{Metric learning objective.} We use the softmax-triplet loss for learning discriminative features early in the training. The loss is defined as,

\begin{equation} \label{eq2}
\mathcal{L}_\textit{tri} = \sum_{i=1}^{(K+1)N} \frac{e^{|| \mathcal{F}(x_i).\mathcal{F}(x_{i+})/\tau||} }{ e^{||\mathcal{F}(x_i).\mathcal{F}(x_{i+})/\tau||} + e^{||\mathcal{F}(x_i).\mathcal{F}(x_{i-})/\tau||} }
\end{equation}

where, for any target image $x_i$, $x_{i+}$ is a positive sample taken from the same cluster, and $x_{i-}$ is a negative sample taken from a different cluster (Initialization Stage, Fig. \ref{fig:side2}).
This step helps the network learn discriminative features for each image, $x_i^o$. Also, $\tau$ denotes temperature parameter and $||.||$ implies L2 distance unless specified.

\subsection{Discriminative clustering}
 Fig. \ref{fig:side2} shows the discriminative clustering stage. Although the pose transformed dataset provides a strong supervisory signal, the learned features may not be discriminative enough for accurate re-identification. We next describe the discriminative clustering strategy which strongly boosts the discriminative power of the network.

At this stage, all the pose-transformed samples used in the initialization phase are dropped from the training data, $\mathbb{D}$. This not only reduces the computational overhead while computing clusters during training but also lets the network concentrate on learning features present in the original dataset, $X_o$. We perform hierarchical clustering \cite{hct_cvpr20} to partition the dataset $X_o$ into $\mathcal{C}$ clusters. We also find the centroid, $c_l$ and radius, $r_l$ for each cluster $C_l$ obtained from the clustering algorithm. While a centroid is estimated as the mean of all the features in the cluster, a cluster radius is defined as the Euclidean distance between features of data-sample from the same cluster farthest from its centroid, and that of the centroid itself. Thus, a cluster centroid is computed as $c_l = mean\{\mathcal{F}(x^o_j)\: |\: x^o_j \in C_l\}$ and a cluster radius as $r_l = \max\{\Vert c_l - x_j^o \Vert, \forall x_j^o \in C_l\}$. These estimated cluster parameters are further used in enforcing the \textit{radial distance}, discussed below.

\textbf{Radial distance loss.}
At this stage we want to encourage large inter-cluster distance while maintaining high similarity among features of data points belonging to the same cluster. We propose to maintain a minimum radial distance for each image from all but the parent cluster in the space. The \textit{radial distance} between a cluster $C_l$ and a feature from any sample $x_i^o \in C_k (l \neq k)$ is defined as, $d_i^l = \Vert c_l - \mathcal{F}(x_i^o) \Vert$. To ensure strict separation between any two clusters, this distance must satisfy the following condition: 
$d^l_i \ge r_l + \gamma$, where $\gamma$ is a margin that ensures a minimum expected separation in terms of the radial distance. 

We define the radial distance loss as
\begin{equation} \label{eq3}
\mathcal{L}_\textit{rd} = \sum_{i=1}^{N}\sum_{l \neq k}^\mathcal{C} \max(0,\: r_l + \gamma - d_i^l)
\end{equation}

where $\gamma$ is the minimum radial distance or margin that needs to be maintained. Thus, if the point $\mathcal{F}(x_i^o)$ lies inside the minimum radial distance, i.e. breaches the minimum margin, for any cluster $C_l$, except its own, it is assigned a penalty. Cluster assignments $r_l$ and $c_l$ are calculated once every epoch and assumed to be the same throughout the epoch. This prevents the network from training on adversely changing cluster radii and centroids, and makes the cluster sizes robust to changing features. For instance, the cluster radius, $r_l$ may actually shrink if the corresponding $\mathcal{F}(x_j^o)$ comes closer to the centroid, $c_l$. However, the minimum radial distance will still prevent any point from coming closer to $C_l$ as $r_l$ is kept constant for the epoch. Similarly, the cluster radius may increase if the corresponding point $\mathcal{F}(x_j^o)$ moves away from $c_l$. Even though the outlier point may now flout the minimum radial distance, it may become a part of a cluster it is closer to, at the next assignment. Thus, the cluster assignment for the data points to $C_l$, and the corresponding cluster radius, $r_l$ is updated in the epoch.

On similar lines, triplet loss aims at maximizing the positive sample distance from the negative sample distance, i.e., it works at point level. On the other hand, \textit{sd} loss works at cluster level, where the distance of the centroid (anchor) to the positive sample (radius) does not change throughout the epoch. All samples therefore inside the radius are considered as belonging to the positive class for one learning epoch. Unlike in triplet loss, a slight movement in the feature space, inside the radius, ensures that cluster assignment does not change for a positive sample. A drastic change signifies an incorrect assignment which is corrected during re-assignment at the next epoch. This makes the algorithm robust to changing cluster assignments and cluster sizes.

Finally, the network is trained using the triplet loss, the classification loss, and the radial distancing loss. The overall loss function is thus given by
\begin{equation} \label{eq4}
\mathcal{L} = \mathcal{L_\textit{cls}} + \lambda_{tri}\mathcal{L_\textit{tri}} + \lambda_{rd}\mathcal{L_\textit{rd}}
\end{equation}

where $\lambda_{tri}$ and $\lambda_{rd}$ are balancing hyperparameters. Triplet loss is now computed using positive samples from the same and negative samples from a different cluster while the number of samples for both classification and triplet loss change to $N$.

\begin{table*}[ht]
	\caption{Comparison of our proposed method with state-of-the-art unsupervised person re-ID methods on Market-1501, DukeMTMC-reID and MSMT17. * number taken from \cite{lin2019bottom}.}
	\vspace{2mm}
	\centering
	\setlength\tabcolsep{7.0pt}
	\resizebox{0.98\textwidth}{!}{
	\begin{tabular}{|l|cccc|cccc|cccc|}
	\hline
 		\multirow{2}{*}{Methods} &  \multicolumn{4}{c|}{\makecell{Market-1501}} & \multicolumn{4}{c|}{\makecell{DukeMTMC-reID}}  & \multicolumn{4}{c|}{\makecell{MSMT17}}\\

 		\cline{2-13} & \makecell{R-1} & \makecell{R-5} & \makecell{R-10} & \makecell{mAP} & \makecell{R-1} & \makecell{R-5} & \makecell{R-10} & \makecell{mAP} & \makecell{R-1} & \makecell{R-5} & \makecell{R-10} & \makecell{mAP}\\
 		
		\hline\hline

		LOMO \cite{liao_cvpr15} & 27.2 & 41.6 & 49.1 & 8.0 & 12.3 & 21.3 & 26.6 & 4.8 & - & - & - & - \\

		BOW \cite{zheng2015scalable} & 35.8 & 52.4 & 60.3 & 14.8 & 17.1 & 28.8 & 34.9 & 8.3 & - & - & - & - \\
		
		OIM \cite{xiao_cvpr17}* & 38.0 & 58.0 & 66.3 & 14.0 & 24.5 & 38.8 & 46.0 & 11.3 & - & - & - & - \\

		BUC \cite{lin2019bottom} & 61.9 & 73.5 & 78.2 & 29.6 & 40.4 & 52.5 & 58.2 & 22.1 & - & - & - & - \\
		
		TAUDL \cite{taudl_eccv18} & 63.7 & - & - & 41.2 & 61.7 & - & - & 43.5 & 28.4 & - & - & 12.5 \\
		
		DBC \cite{ding2019dispersion} & 69.2 & 83.0 & 87.8 & 41.3 & 51.5 & 64.6 & 70.1 & 30.0 & - & - & - & - \\
		
		TSSL \cite{wu_aaai20} & 71.2 & - & - & 43.3 & 62.2 & - & - & 38.5 & - & - & - & - \\		

		SSL \cite{lin_cvpr20} & 71.7 & 83.8 & 87.4 & 37.8 & 52.2 & 63.5 & 68.9 & 28.6 & - & - & - & - \\ 
		
		HCT \cite{hct_cvpr20} & 80.0 & 91.6 & 95.2 & 56.4 & 69.6 & 83.4 & 87.4 & 50.7 & - & - & - & - \\
		
		MMCL \cite{wang_cvpr20} & 80.3 & 89.4 & 92.3 & 45.5 & 65.2 & 75.9 & 80.0 & 40.2 & - & - & - & - \\
		
		CycAs \cite{cycas_eccv20} & 84.8 & - & - & 64.8 & 77.9 & - & - & 60.1 & 50.1 & - & - & 26.7 \\
		
		Ge et al. \cite{ge2020selfpaced} & 88.1 & 95.1 & 97 & 73.1 & - & - & - & - & 42.3 & 55.6 & 61.2 & 19.1 \\
		
		CAP \cite{Wang2021camawareproxies} & 91.4 & 96.3 & 97.7 & 79.2 & 81.1 & 89.3 & 91.8 & 67.3 & 67.4 & 78.0 & 81.4 & 36.9 \\
		
		\hline
		
		\textbf{\textit{Ours}} & \textbf{93.6} & \textbf{97.2} & \textbf{98.3} & \textbf{81.6} & \textbf{86.3} & \textbf{92.4} & \textbf{94.7} & \textbf{70.1} & \textbf{69.9} & \textbf{80.3} & \textbf{85.4} & \textbf{40.7}\\
		\hline
	\end{tabular}}
	\label{tab:market_duke}
\end{table*} 

\section{Experiments}
\label{sec:exp}
\subsection{Datasets and Evaluation Metrics} 
Experiments are conducted on three large-scale datasets - Market1501 \cite{zheng2015scalable}, DukeMTMC-ReID and MSMT17 \cite{wei2018person} datasets. The Cumulative Matching Characteristic (CMC) curve and mean Average Precision (mAP) evaluation metrics are reported on these datasets.

\textbf{Market1501} \cite{zheng2015scalable} dataset consists of 32,668 images from 1,501 identities captured using 6 cameras. The training set consists of 12,936 images from 751 identities. The gallery and query set consist of 19,732 images and 3,368 images respectively from 750 identities. 

\textbf{DukeMTMC-ReID} \cite{ristani2016MTMC} \cite{wu_cvpr18} dataset is a subset of the DukeMTMC \cite{ristani2016MTMC} tracking dataset. It consists of 36,411 images from 1,404 identities captured using 8 cameras. The dataset is split into 16,522 images from 702 identities for the training set, 17,661 images from 702 identities for the gallery set, and 2,228 images from 702 identities for the query set. 

\textbf{MSMT17} \cite{wei2018person} dataset consists of 126,411 images from  4,101 identities captured using 15 cameras (12 outdoor and 3 indoor). The training set consists of 32,621 images from  1,041 identities. The gallery and query set consist of 82,161 and 11,659 images respectively.

\subsection{Implementation details}

We adopt ResNet-50 \cite{he2016deep} pre-trained on ImageNet \cite{deng2009imagenet} as the backbone network for a fair comparison with other works.
It is then trained on PT dataset for $40$ epochs using classification and triplet losses. The Adam optimizer is adopted with a weight decay of $0.0005$, and the learning rate is set to $3.5\times10^{-4}$. For each identity from the generated training set, a mini-batch of $64$ is sampled with $P = 16$ randomly selected identities and $M = 4$ randomly sampled images for computing the softmax triplet loss. 
Next, before each epoch, we perform hierarchical clustering \cite{hct_cvpr20} on the extracted features from the initialization stage, followed by the computation of cluster centroids and radii. The labels obtained from the clusters are used to train the model along with the classification, triplet, and radial distancing losses.
We use the PyTorch library for implementation and train the model on a single RTX 2080 GPU.
%
\subsection{Comparison with State-of-the-Arts}
We compare our proposed approach with prior arts including LOMO \cite{liao_cvpr15}, BOW \cite{zheng2015scalable} that are based on hand-crafted features, TAUDL \cite{taudl_eccv18}, TSSL \cite{wu_aaai20}, and CycAs \cite{cycas_eccv20} that use tracklets, among other unsupervised learning methods \cite{wang_cvpr20, lin_cvpr20, ding2019dispersion, lin2019bottom, ge2020selfpaced}.
Table~\ref{tab:market_duke} shows the Rank-1 and mAP results on Market-1501 \cite{zheng2015scalable}, DukeMTMC-reID \cite{zheng_iccv17}, and MSMT17 \cite{wei2018person} datasets.
We achieve state-of-the-art performance on Market1501 dataset with 93.6\% R-1 and 81.6\% mAP, improving on the prior art \cite{Wang2021camawareproxies} by 2.2\% in R-1 and 2.4\% in mAP. Similarly, for the DukeMTMC-reID dataset, we report 86.3\% R-1 and 70.1\% mAP, an improvement of 5.2\% in R-1 and 2.8\% in mAP over \cite{Wang2021camawareproxies}. Our method also outperforms the state-of-the-art \cite{Wang2021camawareproxies} by \textbf{2.5\%} in Rank-1 accuracy and \textbf{3.8\%} in mAP on the more challenging MSMT17 dataset.
%

%

%

\begin{table}[htp]
	\footnotesize
	\centering
    \begin{minipage}{.38\linewidth}
	\caption{ 
	Ablation study comparing traditional augmentation techniques \textit{AUG} with PT dataset and the importance of radial distance loss ($\mathcal{L}_\textit{rd}$) on Market-1501 dataset.
	}
	\vspace{2mm}
	\centering
	\resizebox{1\textwidth}{!}{
	\setlength\tabcolsep{6pt}
	\begin{tabular}{|c|c|c|c|cc|}
	\hline
 		$X_o$ & AUG & PT & $\mathcal{L}_\textit{rd}$ & R-1 & mAP \\

        \hline\hline
        
        \xmark & \xmark & \xmark & \xmark & 7.8 & 2.6 \\
        
        \cmark & \xmark & \xmark & \xmark & 28.2 & 10.5 \\
        
        \cmark & \cmark & \xmark & \xmark & 34.9 & 14.6 \\
        
        \cmark & \cmark & \xmark & \cmark & 42.2 & 19.9 \\
        
        \cmark & \xmark & \cmark & \xmark &  58.7 & 27.7 \\	
        
        \cmark & \xmark & \cmark & \cmark & 90.3 & 76.1 \\
        
        \cmark & \cmark & \cmark & \cmark & 93.6 & 81.6 \\
        
		\hline
	\end{tabular}
	\label{tab:pt_aug}}
	 \end{minipage}
	\quad
     \begin{minipage}{.27\linewidth}
	\caption{ 
	Ablation study showing the effectiveness of the PT dataset when integrated with recent state-of-the-arts for Market-1501 dataset.
	}
	\vspace{2mm}
	\centering
	\resizebox{1\textwidth}{!}{
	\setlength\tabcolsep{3pt}
	\begin{tabular}{|c|cc|}
	\hline
 		Method & R-1 & mAP \\

        \hline\hline
        
        MMCL \cite{wang_cvpr20} &  84.2 & 51.6 \\	
        
        Ge et al. \cite{ge2020selfpaced} & 90.1 & 77.4 \\
        
        CAP \cite{Wang2021camawareproxies} & 92.7 & 80.3 \\
        
		\hline
	\end{tabular}
	\label{tab:pt_aug2}}
	 \end{minipage}
	\quad
     \begin{minipage}{.29\linewidth}
	\caption{ 
	Ablation on using intuitive and easy to obtain single image body parts annotation (SIA) vs body parts cum pose estimation (PE \cite{kolotouros2019spin}) method on Market-1501.
	}
	\vspace{2mm}
	\centering
	\resizebox{1\textwidth}{!}{
	\setlength\tabcolsep{4pt}
	\begin{tabular}{|c|c|c|cc|}
	\hline
 		SIA & PE & $\mathcal{L}_\textit{rd}$ & R1 & mAP \\

        \hline\hline
        
        \cmark & \xmark & \xmark &  59.1 & 28.5 \\	
        
        \xmark & \cmark & \xmark &  58.7 & 27.7 \\
        
        \cmark & \xmark & \cmark &  90.1 & 76.8 \\
        
        \xmark & \cmark & \cmark &  90.3 & 76.1 \\
        
		\hline
	\end{tabular}
	\label{tab:pt_aug3}}
	 \end{minipage}
\end{table} 

\subsection{Ablation Study}
\label{sec:ablation}

\textbf{Effectiveness of PT dataset.} We perform an extensive evaluation on Market-1501 dataset under different settings as shown in Table~\ref{tab:pt_aug}. We refer to traditional image-level augmentation techniques such as flip, tilt, jitter, random crop etc. used in re-ID methods as \textit{AUG}. The first row shows direct evaluation on ImageNet pre-trained ResNet50 which fails miserably. Second row shows results for training only on original dataset, $X_o$. Next, employing \textit{AUG} provides the network a better discriminative ability. Using PT dataset as a replacement to \textit{AUG} shows a significant boost in performance of the re-ID model. This establishes the effectiveness and importance of using highly diverse and non-trivial augmentations which we perform using pose space transformations. We also demonstrate this in Table~\ref{tab:pt_aug2} by using PT dataset with state-of-the-art methods, reporting an improved performance on the respective original baselines. Lastly, Table~\ref{tab:pt_aug3} compares the performance of our approach when using the two techniques of extracting body parts of a person as described in Sec.~\ref{sec:approach}. One can easily obtain body parts from a single image annotation (SIA) which can then be used for the complete dataset. This method fares favorably to the one used for obtaining both poses and body parts \cite{kolotouros2019spin} but has the advantage of not requiring a body part estimator. One can then freely choose from any of the abundantly available off-the-shelf pose estimation models.

\textbf{Effectiveness of radial distance loss.}
 We hypothesize, a well-learned feature space will learn similar features closer together than dissimilar features. However, in some cases, even the similar features may belong to different identities and therefore must be strictly separable. This would enable the method to distinguish between the small yet relevant differences often existing between similar identities. The rd loss, $\mathcal{L}_\textit{rd}$ maintains a strict separability of clusters owing to the cluster radius and margin distances. This ensures that points from neighboring clusters do not overlap, making them well separable. Furthermore, because the penalty outside the minimum distance is zero, this lets features from similar-looking identities reside close by in the feature space. We illustrate this in the next subsection. Table~\ref{tab:pt_aug}, shows that $\mathcal{L}_\textit{rd}$ enables the model to learn powerful discriminative features on both \textit{AUG} and PT datasets, yielding state-of-the-art results on benchmark datasets. We obtain best results (see last row Table~\ref{tab:pt_aug}) when using augmentations along with our PT dataset for our discriminative learning strategy.
%

\begin{figure}
\centering
\begin{minipage}{.69\textwidth}
  \centering
  \includegraphics[width=1.0\linewidth]{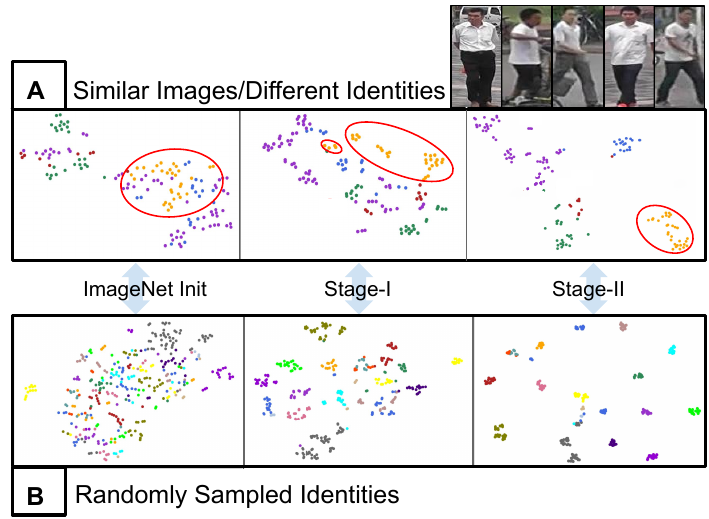}
  \captionof{figure}{t-SNE visualization for 
	\textbf{A.} 5 clusters with similar appearing but different identities, represented by images in the top row.
	\textbf{B.} 20 randomly chosen clusters}
  \label{fig:tsne}
\end{minipage}\enskip\quad
\begin{minipage}{.24\textwidth}
  \centering
  \includegraphics[width=1.0\linewidth]{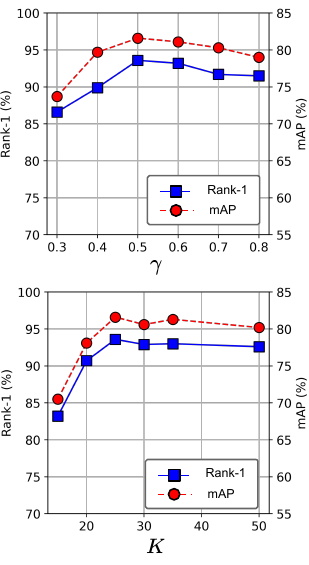}
  \captionof{figure}{Ablation on the Market-1501 by varying the margin $\gamma$, and sample scaling factor $K$.}
  \label{fig:ablation}
\end{minipage}
\end{figure}

\textbf{Effectiveness of our discriminative learning method.} To verify that the proposed method does indeed maximize inter-clusters distances and minimize intra-cluster distances, we plot the t-SNE \cite{Maaten_tsne} visualization shown in Fig.~\ref{fig:tsne}. Panel \textit{A} shows the plot for 5 different identities that appear similar due to their clothing patterns. The first plot represents the clusters when plotted using ImageNet initialized weights.
The highlighted ID in yellow shows that images of the same identity are not separable. The second plot depicts the results from our initialization stage. The network has started to identify the discriminative features. Images from the highlighted ID are seen to have come together, whereas a small number of images of the same ID have clustered at some distance. Finally, the last plot depicts the clusters obtained after training finishes. The clusters have become more distinguishable showing the high discriminative nature of the learned features.
Though the final goal is not to obtain perfect clusters, nevertheless, the model is able to distinguish between different similar looking identities. Similarly, Panel B illustrates results on 20 randomly chosen identities.

\textbf{Analysis on minimum radial distance margin $\gamma$.} The minimum radial distance margin ensures that clusters are at least a minimum distance $\gamma$ from one another. A small margin may lead to images being close in the feature space, and thus being misclassified. A high margin on the other hand may lead to clusters themselves having a large radius. This would mean that images of the same cluster can move far away from each other within the cluster. An optimum margin keeps the clusters compact enough while keeping the adjacent clusters at least preset  distance away. Fig. \ref{fig:ablation} shows the ablation on selecting the best $\gamma$ value.

\textbf{Analysis on sample scaling factor $K$.} It may intuitively seem that a high value of $K$ will give better results. Fig. \ref{fig:ablation} gives the evidence that the performance improvement with increasing $K$ is only up to a certain value and post that the performance stagnates. We provide a simple reason for this. PT dataset is created using randomly sampled poses which are all plausible. However, a pose estimation model may not always be effective against occlusion, missing body parts or background clutter. This leads to inaccurate pose and body part detection. Any random pose transformation applied to this image will become contaminated with source errors, leading to the generation of noisy samples.

\section{Conclusion}
We present an expertly designed, identity independent pose-transformed dataset. This constitutes spatially transforming images in the pose space for generating a realistic and diverse set of images. We show this can be easily adopted to other person re-ID methods. Furthermore, the novel radial distancing criterion introduced attends to the fundamental goal of feature learning, i.e., low intra-cluster and high inter-cluster variation, thereby learning strong discriminative features as established by extensive experimental analysis.

\section{Acknowledgements}

This work is supported by a Start-up Research Grant (SRG) from SERB, DST, India (Project file number: SRG/2019/001938). We thank Pradyumna YM for helping with the code for PT dataset generation.

\bibliography{egbib}
\end{document}